\documentclass{article}
\usepackage{graphicx}
\usepackage{amsmath}

 \usepackage[final]{neurips_2025}

\usepackage[utf8]{inputenc} % allow utf-8 input
\usepackage[T1]{fontenc}    % use 8-bit T1 fonts
\usepackage{hyperref}       % hyperlinks
\usepackage{url}            % simple URL typesetting
\usepackage{booktabs}       % professional-quality tables
\usepackage{amsfonts}       % blackboard math symbols
\usepackage{nicefrac}       % compact symbols for 1/2, etc.
\usepackage{microtype}      % microtypography
\usepackage{xcolor}         % colors
\usepackage[most]{tcolorbox}
\usepackage{lipsum} % For dummy text, you can remove this
\usepackage{wrapfig}
\newtcolorbox{promptbox}[2][]{
    colback=black!5!white,
    colframe=black!75!black,
    fonttitle=\bfseries,
    coltitle=black,
    title=#2,
    breakable,
    enhanced,
    attach boxed title to top left={xshift=10mm, yshift=-2.5mm},
    boxed title style={
        colback=white,
        frame code={
            \path[draw=black!75!black,line width=0.5mm]
            ([xshift=-4mm]frame.north west) -- (frame.north east);
        }
    },
    #1
}

\title{To See or To Read: User Behavior Reasoning in Multimodal LLMs}
\author{Tianning Dong\thanks{Equal contribution.} \quad Luyi Ma$^{*}$ \quad Varun Vasudevan \quad Jason Cho \quad Sushant Kumar\\
\textbf{Kannan Achan}\\
Personalization Team, Walmart Global Tech\\
Sunnyvale, California, USA \\
\texttt{\{tianning.dong, luyi.ma, varun.vasudevan, jason.cho, sushant.kumar,}\\
\texttt{kannan.achan\}@walmart.com}\\
}

\begin{document}

\maketitle

\begin{abstract}

Multimodal Large Language Models (MLLMs) are reshaping how modern agentic systems reason over sequential user-behavior data. However, whether textual or image representations of user behavior data are more effective for maximizing MLLM performance remains underexplored. We present \texttt{BehaviorLens}, a systematic benchmarking framework for assessing modality trade-offs in user-behavior reasoning across six MLLMs by representing transaction data as (1) a text paragraph, (2) a scatter plot, and (3) a flowchart. Using a real-world purchase-sequence dataset, we find that when data is represented as images, MLLMs next-purchase prediction accuracy is improved by 87.5\% compared with an equivalent textual representation without any additional computational cost.

\end{abstract}

\section{Introduction}
\begin{figure}[h]
    \centering
    \includegraphics[width=1.0\linewidth]{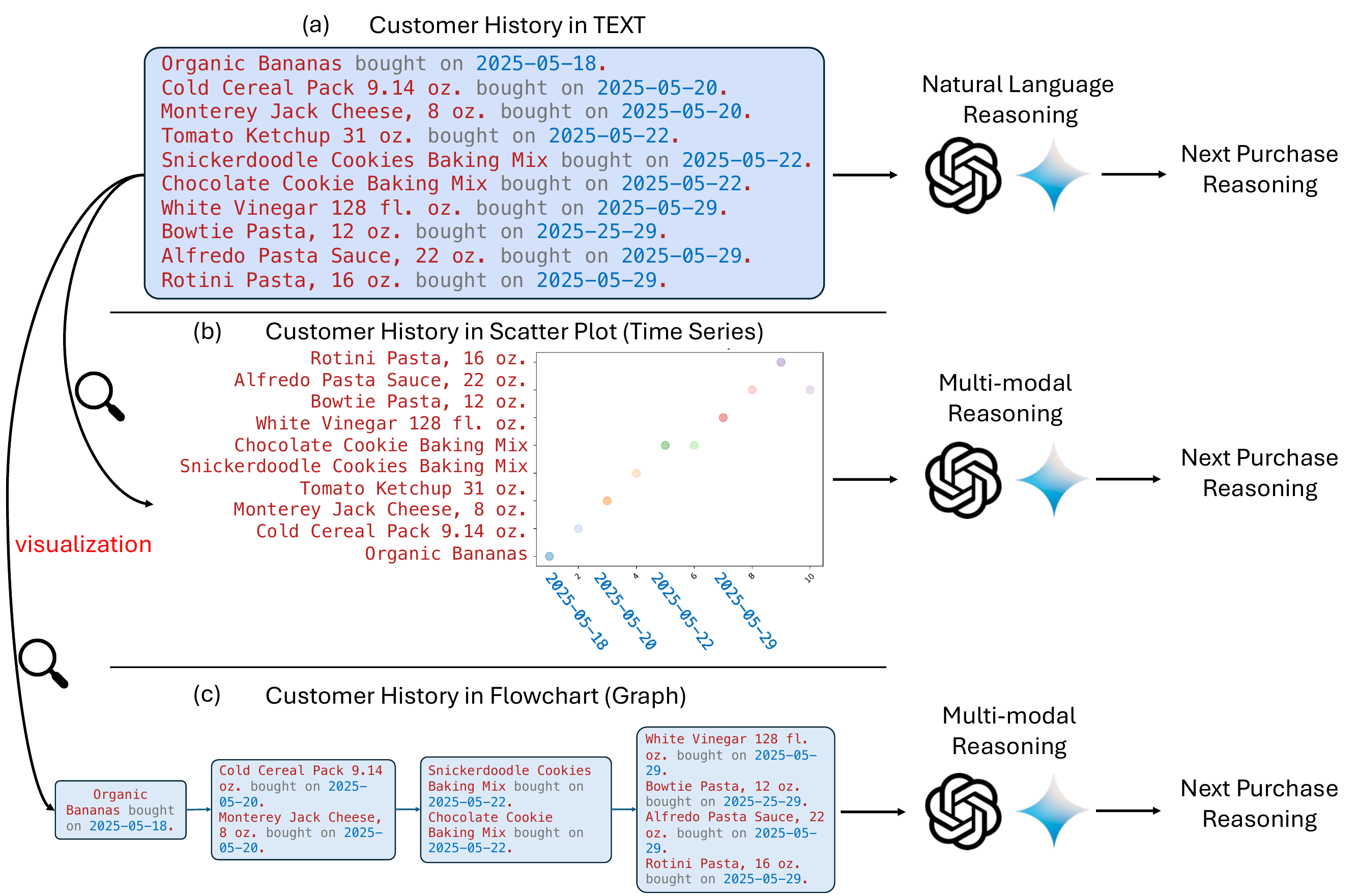}
    \caption{A schematic showing different representations (text, scatter plot, and flow chart) of user history for next purchase prediction using multimodal LLMs.}
    \label{fig:motivation}
\end{figure}

The ability to reason about user behavior from sequential data is a cornerstone of modern intelligent systems, enabling applications such as personalized recommendation (transaction history), chatbot (dialogue history) and proactive user support (meeting notes) \citep{ma2025grace, zhao2024recommender, singh2024historical}. Advancement in Multimodal Large Language Models (MLLMs) has fundamentally shifted the paradigm for this task, replacing specialized models with multimodal general-purpose reasoning engines \citep{li2024llms, shahriar2024putting}. 

This new paradigm, however, introduces a critical and under-explored question: \textbf{how should sequential user histories be represented to optimize for both reasoning accuracy and computational efficiency} \citep{ai2024advancement, li2025graph, imam2025can, yang2025benchmarking}? This question is particularly critical in agentic recommendation systems \citep{huang2025towards, shang2025agentrecbench}, where understanding the user journey and their transition directly impacts personalization and revenue.

The complex user journey with a high-dimensional sequence of clicks, views, and purchases provides a representative testbed for this challenge via next purchase prediction \citep{chen2022sequential}. Effective reasoning requires capturing latent intent and its transition from these sequences, such as distinguishing ``goal-oriented price comparison" from ``aimless browsing" \citep{bauboniene2015commerce, zeng2020user}. The standard approach of feeding MLLMs a flattened, line-by-line textual transcript of these events preserves granular detail, but it losses the structural information of user journey such as topological pattern. Consequently, this method can be inefficient and fail to grasp the holistic user narrative, degrading the quality of intent prediction.

To address this question, we introduce \texttt{BehaviorLens}, a systematic benchmarking framework for evaluating modality trade-offs in user behavior reasoning. \texttt{BehaviorLens} directly compares two representations for the same user history: (i) a high-fidelity textual transcript and (ii) a holistic visual rendering. We evaluate these using state-of-the-art MLLMs on a next-purchase prediction task with a fixed candidate set, assessing both the prediction and the computational cost. Applied to real-world user transaction data, \texttt{BehaviorLens} reveals a key finding: holistic image representations of user history can help improve next-purchase prediction accuracy when compared to text representations, with no additional computational cost.

By systematically comparing modalities under controlled conditions, \texttt{BehaviorLens} contributes both a reproducible methodology for evaluating multimodal reasoning and empirical insights into the efficiency-performance trade-offs of input representations. Beyond next purchase prediction, our framework provides a template for optimizing user historical data streams for MLLMs in any domain involving complex sequential data, supporting future research on efficient context management, data compression, and the development of more robust reasoning systems.

\section{Methodology}   
An agentic recommendation system relies on MLLM's reasoning of user history to understand the user intent. Formally, the reasoning process could be defined as an optimization of the recommendation policy $\pi(a)$ given the user $u$'s history,
\begin{equation}
    \text{max}_{\pi(a)} \mathbb{E} [\mathcal{R}_{MLLM}(u, [\phi(a, i, e)]_n)) | \pi(a)],
\end{equation}
where $\mathcal{R}$ refers to the utility score of user reasoning based on the last $n$ interactions $[\phi(a, i, e)]_n$. Each interaction is defined by a set of actions $a \in A$ (e.g., purchase), a set of engaged items $i \in \mathcal{I}$, and the environment context $e \in \mathcal{E}$ (e.g., timestamp).
We study the impact of user behavior representation function $\phi(a, i, e)$ on user intent reasoning efficiency $\mathcal{R}$.

\subsection{User Behavior Representations}
We compare three ways of representing the user journey as MLLM input with the focus on the purchase history, $a = \text{``purchase''}$.  

\noindent \textbf{Textual Sequential Representation}: the widely-adopted solution for user modeling \citep{liao2023llara} where each tuple $(a, i, e)$ is represented as a natural language description 
\begin{equation}
    \phi_{text}(a, i, e) = \text{``\textit{item \{i\} was \{a\} at timestamp \{e\}}''},
\end{equation}
and the purchase history $[\phi(a, i, e)]_n$ is a concatenation of all the purchase event description.

\noindent \textbf{Scatter-plot Representation}: The same purchase history is transformed into a scatter plot image to visualize the temporal patterns as a time series. It is inspired by visualization in time-series modeling tasks \citep{wang2015image}, 
\begin{equation}
    \phi_{scatter-plot}(a, i, e) = \text{plot}(a, i | x = r(e), y=r(i)).
\end{equation}
Here, $r(\cdot)$ denotes a transformation function to map the value into the coordinate system. For example, $r(\cdot)$ could be a ranking function, mapping each input element to its corresponding ordinal rank. 

\noindent \textbf{Flow-chart Representation}: Inspired by recent work that claims visual compression aids LLM reasoning in structured tasks \citep{li2025graph}, the same purchase history is transformed into a flowchart where purchases as nodes are connected to keep temporal proximity. Typically, we recursively define the flowchart conversion as
\begin{equation}
    \phi_{flowchart}(a, i, e) = \text{node}(a, i, e | p=\{\text{node}(a_{-1}, i_{-1}, e_{-1})\}, s=\{\text{node}(a_{+1}, i_{+1}, e_{+1})\}),
\end{equation}
where the predecessor node set $p$ and the successor node set $s$ only contains the interaction ranked chronologically before ($e_{-1})$ and after ($e_{+1}$) the current interaction $(a, i, e)$, respectively.

\subsection{User Intent Prediction and Reasoning}
For each user, an MLLM receives either the text or the image representation, and is asked to: (1) predict the most likely next purchase $\pi(a)$ from a recall set, (2) provide explanation for the prediction. 
For each representation, we compare the prediction $\pi(a)$ with the ground-truth user behavior, measure the computational cost using token count as a proxy, and evaluate the generated explanation using LLM-as-a-Judge.

\section{Experiments}

In this section we summarize which representation offers the best balance between reasoning accuracy and computational efficiency. See \autoref{sec:dataset_and_metrics} for details on the dataset and the experimental setup and \autoref{sec:prompts-reasoning} for the full prompts used in the experiments.

\begin{table}[!h]
\centering
\caption{Comparison of MLLM reasoning performance with text vs image inputs.}
\begin{tabular}{lccccc}
\toprule
LLM Model & Input Type &  Accuracy  &  Similarity & Token Count & Latency(second)\\
\midrule
 Gemini-2.0-flash-lite & Text & 0.260 & 0.500 & 1233.46 & 1.510\\
Gemini-2.0-flash-lite & Scatterplot & \textbf{0.270} & \textbf{0.528} & 3560.49 & 2.338\\
 Gemini-2.0-flash-lite & Flowchart  & 0.260 & 0.517 & 1525.07 & 1.993\\
\midrule 
 Gemini-2.0-flash & Text & 0.240 & 0.483 & 1228.85 & 1.444\\
 Gemini-2.0-flash & Scatterplot & 0.270 & 0.510 & 3596.50 & 4.580\\
 Gemini-2.0-flash & Flowchart & \textbf{0.310} & \textbf{0.526} & 1529.35 & 4.977\\
\midrule 
 Gemini-2.5-flash-lite & Text & 0.360 & 0.570 & 1219.72 & 1.444\\
 Gemini-2.5-flash-lite & Scatterplot & \textbf{0.530} & \textbf{0.689} & 3623.25 & 2.057\\
 Gemini-2.5-flash-lite & Flowchart & 0.300 & 0.530 & 1566.41 & 1.814\\
\midrule 
 Gemini-2.5-flash & Text & \textbf{0.260} & \textbf{0.479} & 3585.01 & 1.444\\
 Gemini-2.5-flash & Scatterplot & 0.210 & 0.471 & 7223.44 & 6.003\\
 Gemini-2.5-flash & Flowchart & 0.220 & 0.447 & 5281.78 & 4.966\\
\midrule 

 GPT-4o & Text & 0.420 & 0.602 & 1105.52 & 5.451\\
 GPT-4o & Scatterplot & \textbf{0.560} & \textbf{0.713} & 1169.48 & 8.954\\
 GPT-4o & Flowchart & 0.300 & 0.527 & 1042.78 & 7.140\\
\midrule 
 GPT-4.1-mini & Text & 0.320 & 0.542 & 1105.27 & 4.680\\
 GPT-4.1-mini & Scatterplot & \textbf{0.600} & \textbf{0.726} & 1039.24 & 7.849\\
 GPT-4.1-mini & Flowchart & 0.340 & 0.563 & 862.07 & 6.051\\

\bottomrule
\end{tabular}
 \label{tab:llm_reasoning_metrics}
\end{table}

\subsection{Results}

Table \ref{tab:llm_reasoning_metrics} shows that representing customer journeys as images helps improve the accuracy, and the similarity score (i.e. cosine similarity of prediction and ground truth in text embeddings). With all the MLLMs, except Gemini-2.5-flash, either scatterplot or flowchart representation results in better prediction than the text representation. Image input achieves at most 33.9\% improvement in similarity score and 87.5\% improvement in prediction accuracy. This improvement is observed in both larger models, such as GPT-4o, and smaller models, such as GPT-4.1-mini and Gemini-2.5-flash-lite \citep{hurst2024gpt, comanici2025gemini}. The structure of image representations could further make a significant difference in accuracy. For flowchart, it shows better accuracy and similarity especially for Gemini-2.0-flash and Gemini-2.5-flash, while other MLLMs prefer scatterplot.

Table \ref{tab:llm_reasoning_metrics} also shows the computational cost in terms of the total number of tokens used by each MLLM. In Gemini models, flowchart and text input have similar token usage while scatter plots require more tokens. In GPT models, token usage remains at the same level for different representations, while accuracy is significantly higher for image representations. Table \ref{tab:llm_reasoning_metrics} further illustrates the latency as the request time in seconds for each MLLM for reference.

\begin{figure}[!h]
    \centering
    \includegraphics[width=\linewidth]{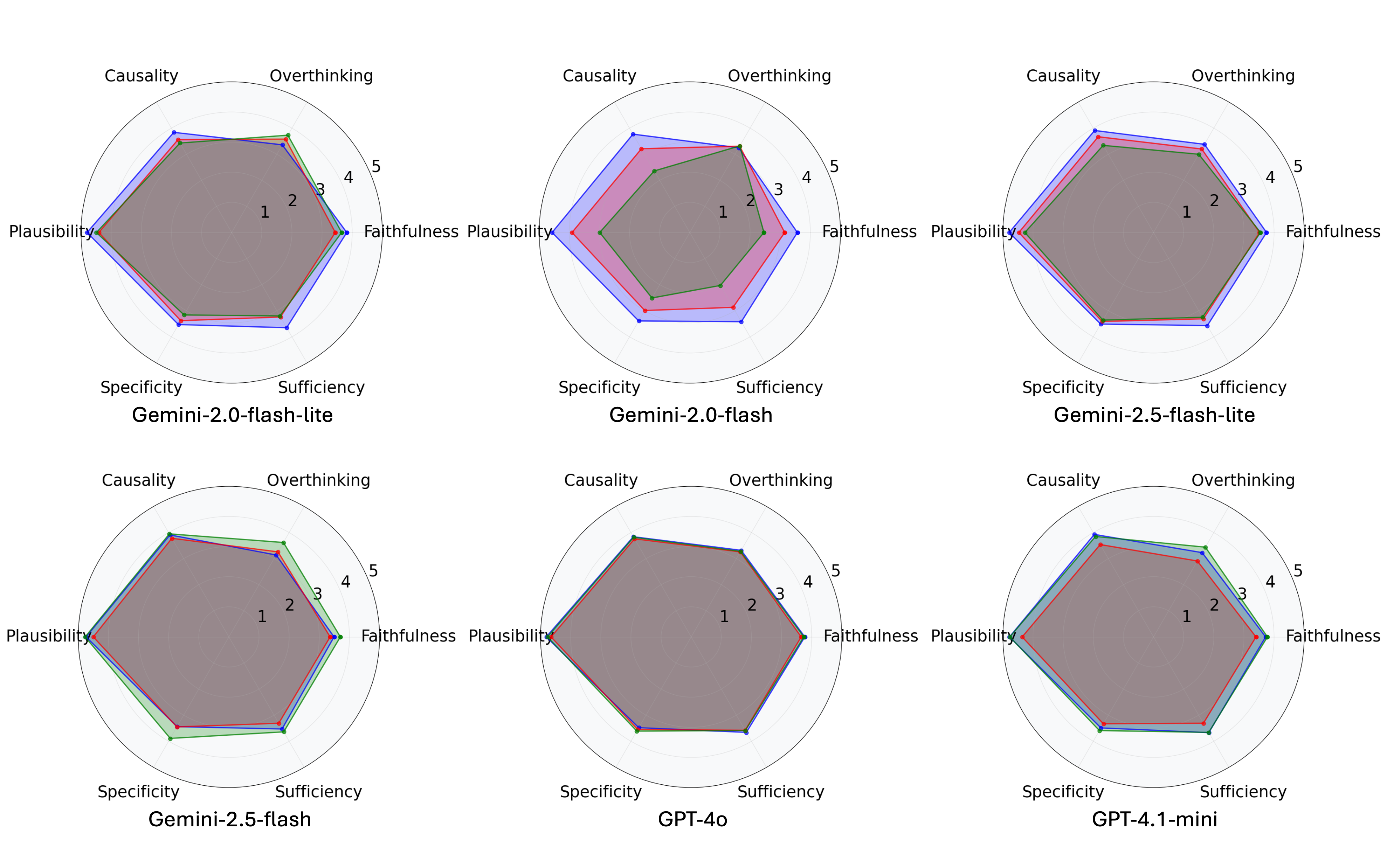}
    \caption{Comparison in MMLM explanations. Each metric is measured by a scale of 1 to 5. Higher score indicates better reasoning in given metric. Blue dots are reasoning of text input. Green dots are reasoning of scatterplot input. Red dots are reasoning of flowchart input.}
    \label{fig:reasoning}
\end{figure}

To better understand whether the improvement in next purchase prediction is impacted by input representations or also impacted by intermediate explanations provided by MLLMs based on understanding of input, we define six metrics to evaluate the quality of explanations. The metrics are faithfulness, overthinking score, causality, sufficiency, specificity, and plausibility. We use LLM-as-a-judge for the evaluation \citep{gu2025surveyllmasajudge}. Prompts used by the judge are provided in \autoref{sec:prompts-eval}. Results are summarized in Figure \ref{fig:reasoning}, which show that the quality of explanations on understanding input representations is not significantly different across different inputs, except Gemini-2.0-flash model. This suggests that the choice of user history representation is the primary driver of improvement in next purchase prediction.

%\lipsum[1-2]

\subsection{Case Study}
Given the sample user prompts in \autoref{sec:prompts-reasoning}, take the output from Gemini-2.5-flash as an example to understand difference in reasoning and prediction for different inputs. The sample customer actually purchased items from Crackers and Granola Bars after the most recent 20 purchases provided as input. 

In the reasoning by text input, it focuses more on the frequency of purchase patterns to predict next item as Cola, while also mentions different possibilities as snacks, which are close to user's next choice in reality. In the reasoning by flowchart input, it focuses more on the recency of purchase and considers the complementary item for the last purchased item. In the reasoning of scatterplot input, it better captures the recurring pattern of purchase, which is unique compared to the other two inputs, and leads to the correct prediction matching with user's interest.

\begin{promptbox}{Output from Text Input}
\textbf{PREDICTION:}
Cola.

\textbf{REASONING:}
The customer purchased Coca-Cola Classic Soda Pop Fridge Pack on both 2019-10-26 and 2019-11-01. This demonstrates a consistent and recurring need for this specific product type, making it highly probable for a subsequent purchase. While other options like Pastries (given the purchase of Toaster Pastries) or other snack/breakfast items are plausible, the direct repeat purchase of Cola makes it the strongest prediction.
\end{promptbox}

\begin{promptbox}{Output from Flowchart Input}
\textbf{PREDICTION:}
Pastries.

\textbf{REASONING:}
The last recorded purchase (indicated by arrow 19) was 100\% Arabica Medium Roast Ground Coffee Pods. Coffee is frequently consumed with breakfast items or snacks. Looking at the user's past purchases, they have previously bought Frosted Toaster Pastries (transaction 14), which falls under the Toaster Pastries product type. This indicates a preference for pastry-like items. Therefore, Pastries is a highly probable next purchase as it complements coffee and aligns with the user's established buying habits for breakfast/snack items.
\end{promptbox}

\begin{promptbox}{Output from Scatterplot Input}
\textbf{PREDICTION:}
Granola Bars.

\textbf{REASONING:}
The user's purchase history shows a recurring pattern of buying Multi Pack Snacks. The most recent purchase (order 20) was Classic Potato Chips, which falls under the Multi Pack Snacks category. Prior to that, OREO Cookie Sticks 'N Creme Dip Snack Packs (order 12) was also purchased from this category. Granola Bars are a very common and popular type of snack item, often sold in multi-packs, making them a highly plausible next purchase to replenish or vary the user's snack supply, directly continuing a recent and established purchasing trend.
\end{promptbox}

\section{Conclusion}

In this paper we investigate the efficiency of reasoning over sequential user behavior data using different representations. We observe significant improvement in prediction accuracy with image representations. This finding advances our understanding of how MLLMs interpret the user journey and holds promise for enhancing personalization in agentic recommendation systems. Future work can focus on further optimizing the visual representations of the user journey to better capture temporal and spatial dynamics in the behavior. Future work should also investigate the impact of longer customer behavior sequences and should benchmark the study on a larger dataset.

\newpage

\bibliographystyle{unsrtnat}
\bibliography{ref}

\newpage
\appendix

\section{Dataset Preparation and Metrics}
\label{sec:dataset_and_metrics}
\subsection{Dataset}
\label{subsec:dataset}
To assess the user behavior reasoning, we sample real-world user shopping interaction from one of the largest e-commerce platform in the United States. We design a benchmark dataset for next-purchase prediction from user purchase histories. 

\noindent \textbf{User Historical Data}: we use the last $n=20$ purchases as the historical data for intent reasoning and prediction. To better focus on the intent prediction and reasoning, we choose the product type of the next purchased items as the ground truth data. The statistics of the benchmark dataset are as shown in Table \ref{tab:stats_dataset}.

\begin{table}[!h]
\centering
\caption{Basic statistics of experiment data}
\begin{tabular}{cc}
    \toprule
    Description & Count \\
    \midrule
    Number of users        & 100  \\
    Number of items        & 1537 \\
    Number of product types & 268 \\
    \bottomrule
\end{tabular}
\label{tab:stats_dataset}
\end{table}

For text-based representation, we provide the MLLM with:
\begin{itemize}
    \item \textbf{Text input}: A sequential description of the user’s purchase history including product names, product types and purchase time in the order they were bought.
    \item \textbf{Multiple-choice candidates}: A set of 20 product types from which the model must select the next likely purchase. Among these 20 product types, 2 product types are the ground truth of user's next purchases while the other 18 product types are randomly sampled from product types that have not been purchased by the customer. 
\end{itemize}

For image-based representation, we provide the MLLM with:
\begin{itemize}
    \item \textbf{Image input}: A flowchart or scatterplot representation of the same purchase history, visually encoding product types and temporal order.
    \item \textbf{Multiple-choice candidates}: A set of 20 product types from which the model must select the next likely purchase. The product types are consistent with multiple choice candidates for text input.
\end{itemize}
 
The LLM is asked to produce two outputs:
\begin{itemize}
    \item \textbf{Prediction}: a predicted next product type from the candidates,
    \item \textbf{Reasoning}: a reasoning explanation describing how the input led to its choice.
\end{itemize}
For evaluation purpose, we also maintain the ground truth of customer's actual next purchase as labels.

\subsection{Metrics and Evaluation}
\label{subsec:metrics_and_eval}
We evaluate results from two perspectives: 1) prediction, 2) reasoning. 

\noindent \textbf{Prediction accuracy}: whether the predicted product type matches a ground-truth products. 
\begin{itemize}
    \item Prediction accuracy: the percentage of users who have ground truth of next-item purchases as an exact match of MLLM predictions.
    \item Similarity score: the maximum of cosine similarities between product types in ground truth and product types in MLLM predictions.
\end{itemize}

\noindent \textbf{Reasoning Evaluation}: 
\begin{itemize}
    \item Faithfulness (1-5): How accurately does the reasoning reflect customer purchase history? 
    \item Overthinking Score (1-5): How well does the reasoning avoid mentioning irrelevant information? 
    \item Causality (1-5): How well does the reasoning present a well-structured argument, or just a list of observations? 
    \item Rationale Plausibility (1-5): How is the reasoning logical and easy to understand? 
    \item Rationale Specificity (1-5): How does the reasoning cite specific data points instead of making generic claims? 
    \item Rationale Sufficiency (1-5): How does the reasoning provide enough evidence to be truly persuasive? 
\end{itemize}

\section{Prompts for User Behavior Reasoning}
\label{sec:prompts-reasoning}
This section shows sample prompts used to predict next best item and provide reasoning. The below example of user history is synthetic data. For text input, the user history is provided in a loggings format with three lines per purchase to describe product names, types and timestamps. For image input, the user history is provided in a flowchart or scatterplot with further instructions on how to read the plots.

\begin{promptbox}{System Prompt}
You are a helpful assistant designed to analyze user behavior in e-commerce. Your goal is to predict the user's next action and provide a brief, data-driven reasoning for your prediction based on the provided user history. 
\end{promptbox}

\begin{promptbox}{Prompt for User History in Text}
\textbf{USER HISTORY:}
\begin{verbatim}
- Product name: Steak & Chop Marinade
- Product type: Barbecue Sauce & Marinades
- The customer bought this product on 2019-10-26 at 12:15:27.
- Product name: Italian Style Finely Shredded Cheese
- Product type: Shredded Cheese
- The customer bought this product on 2019-10-26 at 12:16:22.
- Product name: Cubed Colby & Monterey Jack Cheese
- Product type: Cubed & String Cheese
- The customer bought this product on 2019-10-26 at 12:16:34.
- Product name: White Round Top Bread Loaf
- Product type: White Bread
- The customer bought this product on 2019-10-26 at 12:17:39.
- Product name: Beefsteak No Seeds Rye Bread
- Product type: White Bread
- The customer bought this product on 2019-10-26 at 12:17:39.
- Product name: Mandarin Orange Sparkling Water
- Product type: Water Enhancers
- The customer bought this product on 2019-10-26 at 12:17:39.
- Product name: Coca-Cola Classic Soda Pop Fridge Pack
- Product type: Cola
- The customer bought this product on 2019-10-26 at 12:17:39.
- Product name: Cherry Limeade Sparkling Water
- Product type: Water Multipacks
- The customer bought this product on 2019-10-26 at 12:17:58.
- Product name: Glacier Freeze/Cherry/Arctic Blitz Sports Drinks
- Product type: Sports & Vitamin Drinks
- The customer bought this product on 2019-10-26 at 12:17:58.
- Product name: 100% Apple Juice
- Product type: Juice
- The customer bought this product on 2019-10-26 at 12:18:13.
- Product name: OREO Cookie Sticks 'N Creme Dip Snack Packs
- Product type: Multi Pack Snacks
- The customer bought this product on 2019-10-26 at 12:18:55.
- Product name: Original Pork Breakfast Sausage Roll
- Product type: Breakfast Sausage
- The customer bought this product on 2019-10-26 at 12:20:00.
- Product name: Cinnamon Toast Crunch Whole Grain Breakfast Cereal
- Product type: Cold Cereal
- The customer bought this product on 2019-11-01 at 10:50:11.
- Product name: Frosted Toaster Pastries, Cookies and Cream
- Product type: Toaster Pastries
- The customer bought this product on 2019-11-01 at 10:51:31.
- Product name: Extra Virgin Olive Oil
- Product type: Olive Oil
- The customer bought this product on 2019-11-01 at 10:54:06.
- Product name: 2% Reduced Fat Milk, Refrigerated
- Product type: Dairy Milk
- The customer bought this product on 2019-11-01 at 10:54:52.
- Product name: Coca-Cola Classic Soda Pop Fridge Pack
- Product type: Cola
- The customer bought this product on 2019-11-01 at 10:56:45.
- Product name: Glacier Cherry Sports Drinks
- Product type: Sports & Vitamin Drinks
- The customer bought this product on 2019-11-01 at 10:57:00.
- Product name: 100% Arabica Medium Roast Ground Coffee Pods
- Product type: Coffee Pods
- The customer bought this product on 2019-11-01 at 10:57:12.
- Product name: Classic Potato Chips
- Product type: Multi Pack Snacks
- The customer bought this product on 2019-11-01 at 10:57:12.
\end{verbatim}
\end{promptbox}

\begin{promptbox}{Prompt for User History in Flowchart}
\textbf{USER HISTORY:}
Given the below flowchart representing user purchase history, where arrows with numbers indicate the ordering of purchase. 
\end{promptbox}
\begin{figure}[!h]
    \centering
    \includegraphics[width=\linewidth]{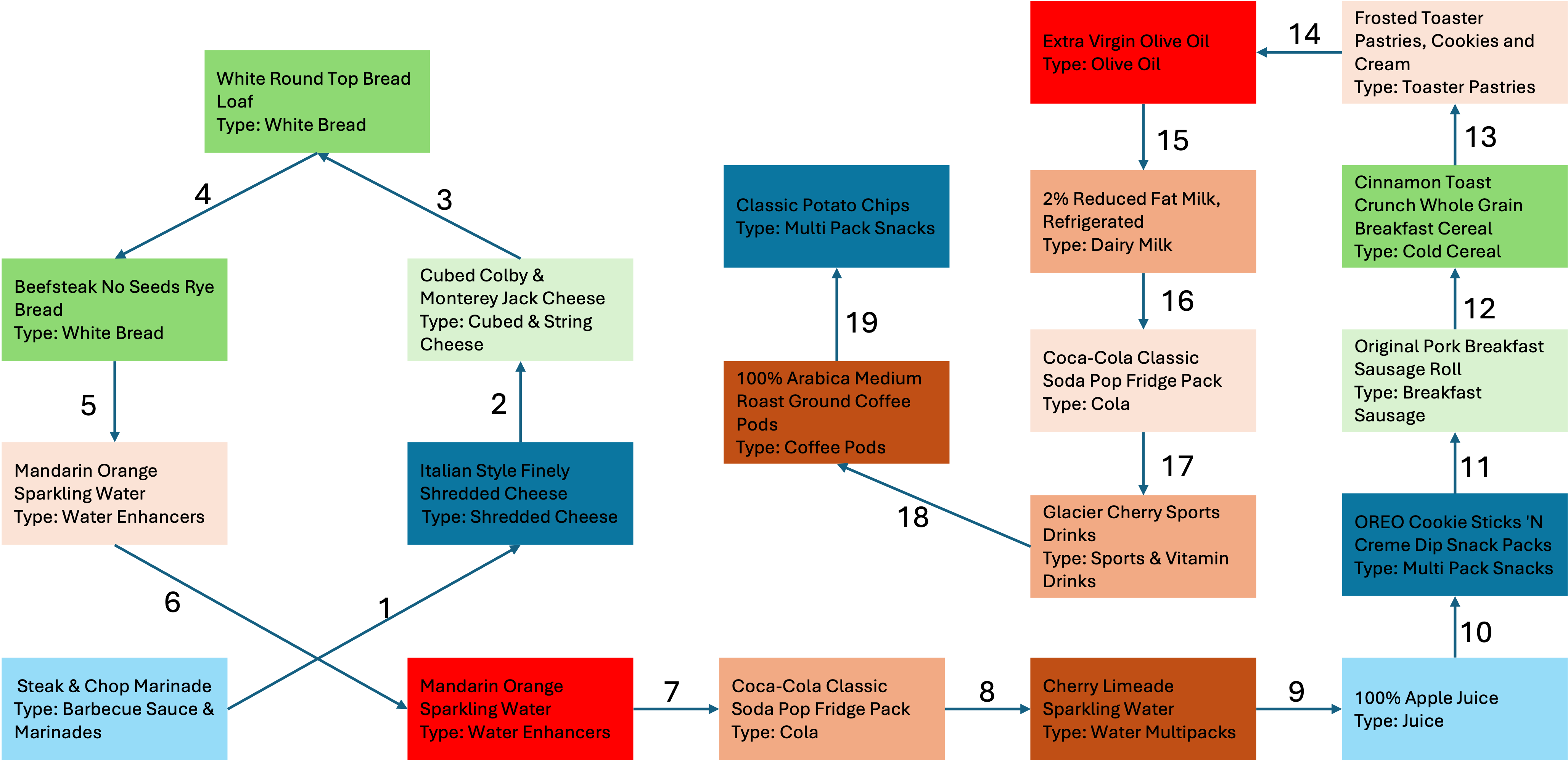}
    \caption{Flowchart of user history}
    \label{fig:scatterplot}
\end{figure}

\begin{promptbox}{Prompt for User History in Scatterplot}
\textbf{USER HISTORY:}
Given the below scatterplot representing user purchase history, where x-axis represents the ordering of purchase, and y-axis represents the product types of purchased items. 
\end{promptbox}
\begin{figure}[!h]
    \centering
    \includegraphics[width=\linewidth]{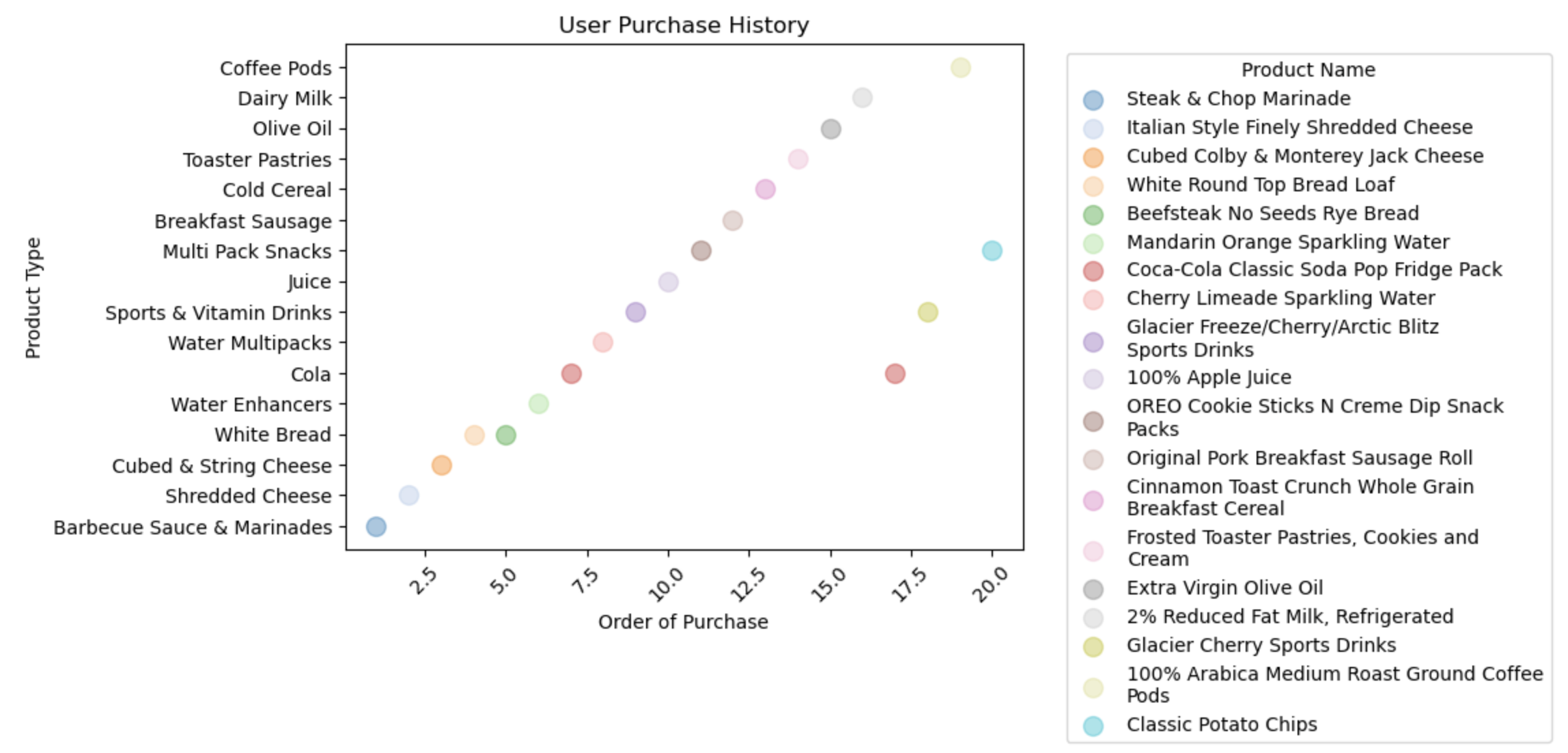}
    \caption{Scatterplot of user history}
    \label{fig:scatterplot}
\end{figure}

\begin{promptbox}{User Prompt}
\textbf{QUESTION:}
Based on the user history provided above, predict what might be next possible purchase and explain why, choosing from the given multiple choices. Provide your answer in a json format with prediction result and reasoning as two keys.

\textbf{CHOICES:}
Granola Bars, Crackers, Instant Coffee, Cola, Whole Fresh Herbs, Chocolate Multipacks, Margarines, Canned Vegetables, Pregnancy and Ovulation Tests, Broth, Stocks and Bouillon, Lunch Packs, Oatmeal and Hot Cereal, Atkins Test, Holiday Dairy and Egg Nogs, Pastries, Support Hose and Socks, Holiday Bakery, Itch and Rash Treatments, Salad Kits and Bowls, Muffins and Scones.
\end{promptbox}

\section{Multi-dimensional Evaluation of Explanation}
\label{sec:prompts-eval}
This section shows sample prompts used to evaluate the explanation provided by MLLMs for next-item prediction, including the details of scoring criteria. To better evaluate the explanation, we need to provide ground truth of user purchase history and actual next purchase for reference.
%{\color{red}{adding prompt template here. Feel free to consider the following [promptbox]}}.

\begin{promptbox}{System Prompt}
You are an expert evaluator of reasoning for next purchase prediction. Your task is to assess the quality of reasoning based on the given specialized criteria.

1. Faithfulness (1-5): How accurately does the reasoning reflect customer purchase history?
\begin{verbatim}
   - 1: Major ground truth errors, contradicts behavior
   - 2: Some ground truth inaccuracies, limited accuracy
   - 3: Generally accurate ground truth with minor issues
   - 4: Highly accurate ground truth representation
   - 5: Perfect ground truth accuracy, nuanced understanding
\end{verbatim}
2. Overthinking Score (1-5): How well does the reasoning avoid mentioning irrelevant information?
\begin{verbatim}
   - 1: Exceptional much irrelevant information  
   - 2: Much irrelevant information  
   - 3: Moderate irrelevant information 
   - 4: Limited irrelevant information 
   - 5: No irrelevant information 
\end{verbatim}
3. Causality (1-5): How well does the reasoning present a well-structured argument, or just a list of observations?
\begin{verbatim}
   - 1: No causality, just a list of observations
   - 2: Limited causality, basic argument
   - 3: Moderate causality, some reasonable argument
   - 4: Good causality, clear argument
   - 5: Exceptional causality, well-structured argument
\end{verbatim}
4. Rationale Plausibility (1-5): How is the reasoning logical and easy to understand?
\begin{verbatim}
   - 1: No irrelevant information
   - 2: Limited irrelevant information
   - 3: Moderate irrelevant information
   - 4: Much irrelevant information
   - 5: Exceptional much irrelevant information
\end{verbatim}
5. Rationale Specificity (1-5): How does the reasoning cite specific data points instead of making generic claims?
\begin{verbatim}
   - 1: No specific information
   - 2: Limited specific information
   - 3: Moderate specific information
   - 4: Much specific information
   - 5: Exceptional much specific information
\end{verbatim}
6. Rationale Sufficiency (1-5): How does the reasoning provide enough evidence to be truly persuasive?
\begin{verbatim}
   - 1: No evidence, not persuasive
   - 2: Limited evidence, less persuasive
   - 3: Moderate evidence, somehow persuasive
   - 4: Enough evidence, well persuasive
   - 5: Exceptional evidence, very persuasive
\end{verbatim}
\end{promptbox}

\begin{promptbox}{User Prompt}
\textbf{User History and Ground Truth}
The below is the list of product types in user purchase history. (Add purchased product types in time order). 
The user actually purchased (add the purchased product types as ground truth). 

\textbf{Reasoning:}
Meanwhile the agent predicted the user would have bought (add the predicted next-purchase product type).
The below is the reasoning of next purchase prediction by the agent. (Add the reasoning provided by MLLMs).

\textbf{Questions:}
Based on reasoning and prediction provided above, generate scores using the above criteria.
\end{promptbox}

\section{Related Works}

\subsection{Next Item Prediction}
Classical approaches to next item prediction include Markov chains, matrix factorization, and factorization machines, which model user–item interactions using latent factors. With the rise of deep learning, recurrent neural networks such as GRU4Rec \citep{hidasi2016sessionbasedrecommendationsrecurrentneural} became influential in modeling sequential recommendation, capturing order-dependent behaviors in clickstreams and purchase sequences. Transformer-based architectures, such as SASRec \citep{kang2018selfattentivesequentialrecommendation} and BERT4Rec \citep{sun2019bert4recsequentialrecommendationbidirectional}, further advanced next-item prediction by leveraging self-attention to model long-range dependencies in user behavior.

Beyond sequential modeling, graph-based approaches such as SR-GNN \citep{10.1609/aaai.v33i01.3301346} and GCE-GNN \citep{10.1145/3397271.3401142} introduced graph neural networks for capturing complex item–item relations in session-based recommendation. Other notable advances include Caser \citep{10.1145/3159652.3159656}, which treats user-item interactions as sequences of images, and STAMP \citep{10.1145/3219819.3219950}, which focuses on capturing users’ short-term interests for better next-item prediction. Graph modality provides extract context for next item prediction with generative models \citep{ma2024triple}.

Despite significant improvements in accuracy, these models have limited interpretability of why a particular item is predicted. This has led to a growing interest in approaches that not only predict but also explain user preferences, paving the way for the integration of large language models into recommendation tasks.

\subsection{LLM for Recommendation and Reasoning}
Large language models (LLMs) such as GPT, LLaMA, and PaLM have recently been explored for recommender systems by framing user histories and items as natural language. Early explorations such as LLMRec \citep{10.1145/3616855.3635853} and ReXPlug \citep{10.1145/3404835.3462939} showed that LLMs can adapt to recommendation tasks with minimal fine-tuning, leveraging their world knowledge to infer user preferences. Other works like P5 proposed a unified framework that casts different recommendation problems into text-to-text tasks, demonstrating the flexibility of LLMs as general-purpose recommenders.

One of the distinctive advantages of LLMs is their ability to generate reasoning alongside predictions. Models such as ChatRec and Explainable GPT-based Recommenders show that natural language explanations can improve user trust and system transparency. Despite these advances, most existing work evaluates LLM reasoning in text-only settings \citep{gao2023chat}. Little is known about how LLMs reason when user histories are compressed into images. Our work is among the first to directly compare the reasoning efficiency of LLMs across textual and image-based representations of customer journeys, expanding the discussion from prediction accuracy to reasoning quality.

\subsection{Multi-Modal Representations of Customer Journey}
Beyond text-based sequence modeling, more work explores the idea of transforming sequential user interactions into image-like representations that can be processed by convolutional or vision-inspired models. For example, Markov Transition Fields (MTF) and Gramian Angular Fields (GAF) transform time series into images, enabling CNNs to extract spatial dependencies that are otherwise hard to capture in sequential models. In the recommender systems domain, Caser explicitly modeled sequential user–item interactions as a 2D ``image” where convolutional filters capture local sequential patterns as well as high-level transition features. Similarly, DeepMove leveraged trajectory data by representing spatio-temporal patterns in image-like forms to predict next locations, which parallels the idea of encoding customer journeys for next-item prediction \citep{zhou2018deepmove}. 

More recent work has used self-attention and visual encodings to compress large-scale sequential behaviors into compact 2D structures. For instance, Time2Graph converts multivariate sequences into graph-like images \citep{cheng2020time2graph}, while Rec2Image proposed encoding user behavior sessions into image matrices, enabling CNNs to identify higher-order dependencies. These studies demonstrate that compressing sequential histories into visual representations not only reduces input complexity but also allows reasoning models to leverage spatial proximity and clustering cues that may be less salient in raw text sequences. 

Despite these advances, prior work primarily used CNN-based architectures for learning from compressed images. Little is known about how large language models (LLMs) reason when fed such image-based summaries compared to traditional text sequences. Our work extends this line of research by investigating reasoning efficiency in LLMs when user journeys are compressed into scatter plot–like images, directly comparing prediction performance and the quality of generated reasoning across modalities.

% \section{Previous discussion}
% A table comparison

% LLava

% One experiment compare different LLMs and see how they perform 
% under different 
% Cambrian, used to understand the visual ability. 

% How to define the front facing? 

% Camera calibration, 

% World Model agent, robot, Physical

% Intro two images comparision

% diffusion multi object generation

% diffusion multi object generation with 3D Perception 

% Contribution:

% 3 objects -- segmentation -- spatial planning 
% -- multi-objects foreground images
% -- Multi-obejct image generation

% First step, multi-objects foreground images

% Diffussion, spatial recognition ability

% 3D recognition ability

% 1. spatial planning

% Existing works, how compared to our works

% Spatial Planning with LLM Agents

% Spatial planning distribution 

% Background already generated

% Add experiments, performance 

% Ground Truth

% We have ground truth, 

% Home, inside we already did the segmentation. 

% How did we get the ground truth data

% Segmentation -- tranining data -- spatial planning training

% Ground truth image, Segmentation, get the foreground images, 

% Use ground truth for training

% 2. image generation

% Diffusion fine-tune, how reasonable it is?

% Novelty, contribution in the model

% with vs without

% Diffusion model, add the lora for better 

% Foreground image generation already exists, 

% Understanding the multiple objects

% No multi-object generation

% product images -VLM-> segmentations -VLM-> resize and positioning -VLM-> composition -VLM-> verify 

\end{document}